\documentclass[letterpaper]{article} 
\usepackage{aaai24}  
\usepackage{times}  
\usepackage{helvet}  
\usepackage{courier}  
\usepackage[hyphens]{url}  
\usepackage{graphicx} 
\usepackage{natbib}  
\usepackage{caption} 
\usepackage{algorithm}
\usepackage{algorithmic}
\usepackage{booktabs}
\usepackage{multirow}
\usepackage{makecell}
\usepackage{amssymb}
\usepackage{amsmath}
\usepackage{newfloat}
\usepackage{listings}
\DeclareCaptionStyle{ruled}{labelfont=normalfont,labelsep=colon,strut=off} 
\floatstyle{ruled}
\newfloat{listing}{tb}{lst}{}
\floatname{listing}{Listing}
\title{Finding Visual Saliency in Continuous Spike Stream}
\author{
    Lin Zhu\textsuperscript{\rm 1}, Xianzhang Chen\textsuperscript{\rm 1}, Xiao Wang\textsuperscript{\rm 3}, Hua Huang\textsuperscript{\rm 2,\rm 1,\thanks{Corresponding author: Hua Huang. }}
}
\affiliations{
    \textsuperscript{\rm 1} School of Computer Science and Technology, Beijing Institute of Technology, China \\
    \textsuperscript{\rm 2} School of Artificial Intelligence, Beijing Normal University, China\\
    \textsuperscript{\rm 3} School of Computer Science and Technology, Anhui University, China
    
      \{linzhu,xianzhangchen\}@bit.edu.cn, 
     wangxiaocvpr@foxmail.com, huahuang@bnu.edu.cn
}

\usepackage{bibentry}

\begin{document}

\maketitle

\begin{abstract}
As a bio-inspired vision sensor, the spike camera emulates the operational principles of the fovea, a compact retinal region, by employing spike discharges to encode the accumulation of per-pixel luminance intensity. Leveraging its high temporal resolution and bio-inspired neuromorphic design, the spike camera holds significant promise for advancing computer vision applications. Saliency detection mimics the behavior of human beings and captures the most salient region from the scenes. In this paper, we investigate the visual saliency in the continuous spike stream for the first time. To effectively process the binary spike stream, we propose a Recurrent Spiking Transformer (RST) framework, which is based on a full spiking neural network. Our framework enables the extraction of spatio-temporal features from the continuous spatio-temporal spike stream while maintaining low power consumption. To facilitate the training and validation of our proposed model, we build a comprehensive real-world spike-based visual saliency dataset, enriched with numerous light conditions. Extensive experiments demonstrate the superior performance of our Recurrent Spiking Transformer framework in comparison to other spike neural network-based methods. Our framework exhibits a substantial margin of improvement in capturing and highlighting visual saliency in the spike stream, which not only provides a new perspective for spike-based saliency segmentation but also shows a new paradigm for full SNN-based transformer models. 
The code and dataset are available at \url{https://github.com/BIT-Vision/SVS}.
\end{abstract}

\section{Introduction}
The human visual system (HVS) possesses an extraordinary ability to swiftly identify and focus on visually distinct and prominent objects or regions within images or scenes \cite{borji2012quantitative}. This remarkable process has inspired advancements in computer vision, particularly in saliency detection, which aims to identify objects or areas of significance carrying valuable information in images or videos \cite{wu2019mutual,fan2019shifting}. As a burgeoning field, saliency detection has attracted the attention of researchers across various disciplines. Central to this pursuit is the detection of salient objects, a process often referred to as saliency detection or salient-object detection. This involves locating and isolating objects from their backgrounds, leading to the development of numerous models that excel in traditional image modalities \cite{wang2021salient}.

However, the sensing mechanism of human vision~\cite{sinha2017cellular} diverges from the standard digital camera paradigm. Human vision lacks the concept of frames or discrete pictures, and its mechanism is considerably intricate.  
Nonetheless, cues and inspiration can be drawn from the structure and signal processing within the human retina. 
Researchers have designed spiking image sensors that mimic the behavior of integrate-and-fire neurons, operating asynchronously~\cite{culurciello2003biomorphic,shoushun2007arbitrated,zhu2019retina,dong2017spike,zhu2020hybrid}. These sensors, in contrast to conventional cameras with fixed integration times, enable each pixel to determine its optimal integration time. Consequently, these spiking image sensors facilitate the reconstruction of visual textures without adhering to the constraints of frames. A recent advancement in this domain is the spike camera \cite{dong2017spike, zhu2019retina}, which adopts a fovea-like sampling method (FSM) and mirrors the structure and functionality of the primate fovea. Unlike dynamic vision sensors based on temporal contrast sampling \cite{lichtsteiner2008128}, the spike camera incorporates spatial (250$\times$400) and temporal (20,000 Hz) resolution, merging visual reconstruction and motion sensitivity to effectively handle high-speed vision tasks.
\begin{figure}[t]
  \centering
  \includegraphics[width=\linewidth]{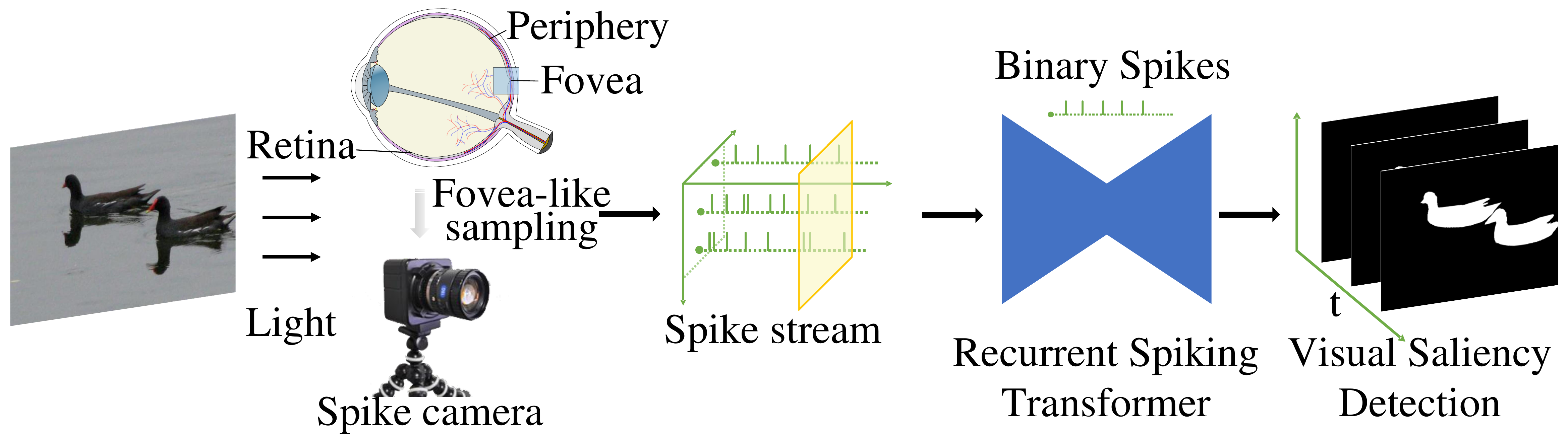}
  \caption{{The motivation of detecting visual saliency in continuous spike stream.} In contrast to ANNs, SNNs provide a biologically realistic model where neurons communicate through discrete spikes, making them well-suited for processing spike data with low power consumption.}
  \label{fig1}
\end{figure}
In this paper, we delve into the field of visual saliency within continuous spike streams. Contrary to traditional image modalities, visual saliency is encoded within binary spike streams in the spatio-temporal domain. 
Given the 20,000 Hz sampling rate of the spike camera, effectively processing the continuous spike stream presents a challenge.
This leads us to a key question: ``\emph{How can visual saliency be detected from a continuous spike stream while minimizing power consumption?}'' 
The potential lies in synergizing continuous spike streams with low-power spiking neural networks (SNNs). Compared to artificial neural networks (ANNs), SNNs offer a more biologically realistic model, with neurons communicating via discrete spikes rather than continuous activation. However, existing SNN researches have predominantly centered on tasks such as classification, optical estimation, motion segmentation, and angular velocity regression, often utilizing traditional or event cameras~\cite{fang2021deep,lee2020spike,zhu2022ultra,zhu2022event}.

To the best of our knowledge, this work pioneers the exploration of visual saliency within continuous spike streams captured by the spike camera. The motivation is shown in Fig. \ref{fig1}. To effectively process binary spike streams, we present the Recurrent Spiking Transformer (RST) framework, a full spiking neural network architecture. Our framework comprises spike-based spatio-temporal feature extraction, recurrent feature aggregation, multi-scale refinement, and multi-step loss. To facilitate model training and validation, we have constructed an extensive real-world spike-based visual saliency dataset, enriched with diverse lighting conditions.
Our contribution can be summarized as follows:

\begin{itemize}
\item We investigate visual saliency within continuous spike streams captured by the spike camera for the first time. To effectively process the binary spike stream, we propose a Recurrent Spiking Transformer (RST) framework, which is based on a full spiking neural network. 

\item We propose a recurrent feature aggregation structure to enhance the temporal property of the spiking transformer. Moreover, a multi-step loss is designed for better utilizing the temporal information of the spike stream.

\item  We build a novel dataset consisting of spike streams and per-object masks. Extensive experimental results on our real-world datasets demonstrate the effectiveness of our network. Our dataset will be available to the research community for further investigation and exploration. 

\end{itemize}

\section{Related Work}
\noindent\textbf{Visual Saliency in Traditional Image}
Salient object detection is an active research field in computer vision, which plays an important role in object segmentation and detection tasks. Depending on the different detection targets, this field can be divided into various sub-tasks. Traditional RGB \cite{wu2019mutual,chen2020global} and RGB-D \cite{ji2021calibrated,fu2021siamese} methods aim to find salient objects from complex scenes through the color and depth information. Co-SOD \cite{zhang2021deepacg,su2023unified} is used to detect the co-saliency objects between a group of images. VSOD \cite{fan2019shifting,yan2019semi,zhang2021dynamic,} pay more attention to using the spatio-temporal feature which is helpful for detecting salient objects in continuous images.

\begin{figure}[t]
    \centering
    \includegraphics[width=\linewidth]{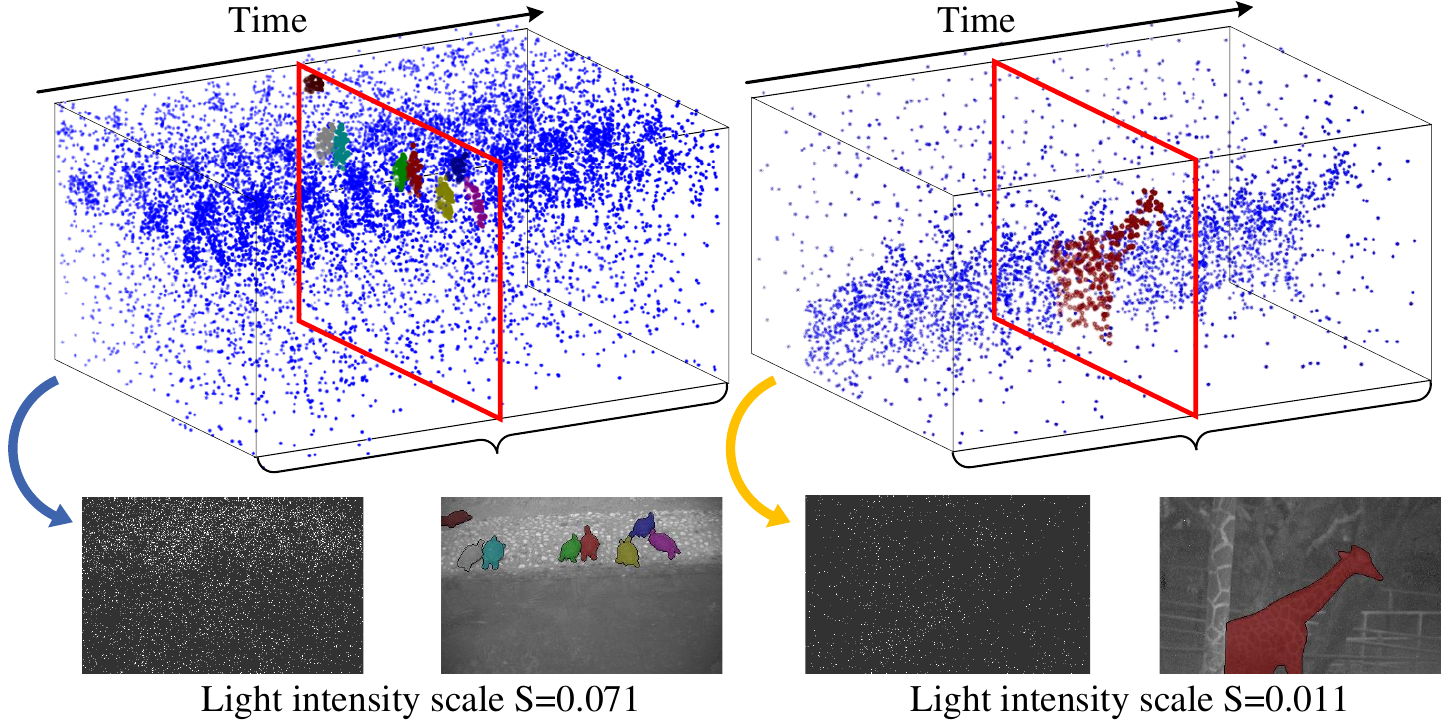}
    \caption{{Visual saliency in spatio-temporal spike stream.}}
    \label{fig2}
\end{figure}

\noindent\textbf{Neuromorphic Camera Applications} Neuromorphic camera, such as Event camera \cite{serrano2013128,brandli2014240} and Spike camera \cite{dong2017spike,zhu2019retina}, which captures the change or accumulation of light intensity, has been widely used in computer vision applications \cite{xiang2021learning,wang2022hardvs,dong2019efficient,gu2023reliable}. For example, E2vid \cite{rebecq2019high} applies ConvLSTM \cite{shi2015convolutional} to extract spatio-temporal features from event streams for video reconstruction. EV-IMO \cite{mitrokhin2019ev} use the continuous property of events to solve motion segmentation task. Spike2Flow \cite{zhao2022learning} and SCFlow \cite{hu2022optical} use spike data to generate optical flow to deal with different speed scenes. RSIR \cite{zhu2023recurrent} is designed to reduce the noise under general illumination for spike-based image reconstruction.

\begin{figure*}[t]
  \centering
  \includegraphics[width=0.8\linewidth]{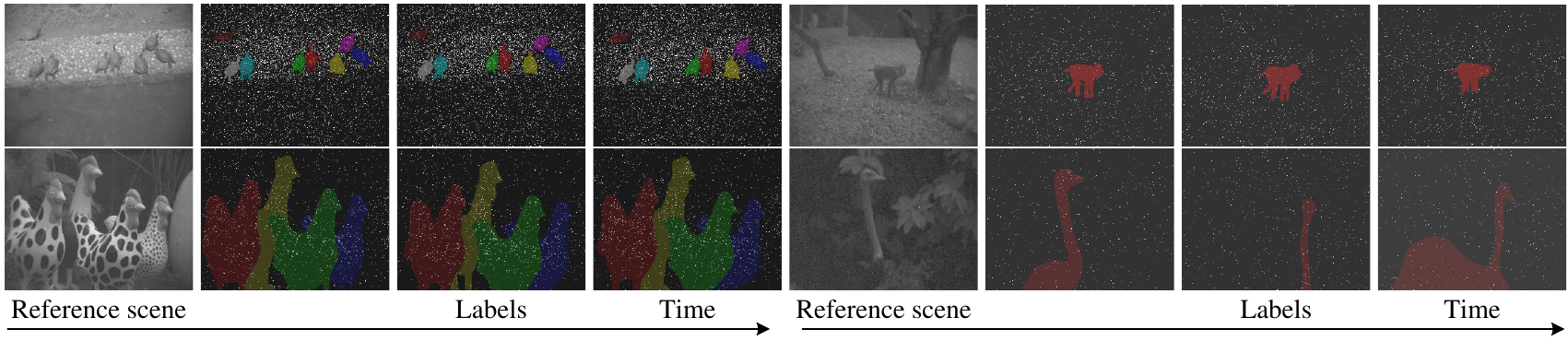}
  \caption{{Samples in our spike-based visual saliency (SVS) dataset.} }
  \label{fig3}
\end{figure*}
\noindent\textbf{Spiking Neural Network for Vision Task}
Based on the capability of simulating neuron dynamics, spiking neural networks have been used for many vision tasks. Spiking-YOLO \cite{kim2020spiking} trains an ANN model and uses multi-step to accumulate spikes for imitating ANN features, which is widely used for SNN training. SEW-ResNet \cite{fang2021deep} and Spikformer \cite{zhou2022spikformer} directly train SNN models for image classification. Spike-Flownet \cite{lee2020spike} and XLIF-FireNet\cite{hagenaars2021self} apply SNN to the optical flow estimation task. Spiking Deeplab \cite{kim2022beyond} is designed to generate a dense prediction for semantic segmentation. EVSNN \cite{zhu2022ultra} uses the temporal information of neuron membrane potential to reconstruct continuous event-based video frames.

Inspired by the temporal property and the low energy consumption of the spiking neural networks, we build a recurrent spiking transformer architecture to extract spatio-temporal spiking-based features, which facilitates the detection of visual saliency in a continuous spike stream.

\section{Visual Saliency in Continuous Spike Stream}
In this section, we first analyze the sampling principle of spike cameras. Based on the characteristics of spike data, we further analyze the visual saliency in spike data and construct a spike-based visual saliency (SVS) dataset.

\noindent\textbf{Spike Sampling Mechanism}
In a spike camera, the photoreceptor converts the intensity of light into voltage \cite{dong2017spike,zhu2019retina}.
When the voltage $I$ surpasses a predetermined threshold $\phi$, a one-bit spike is generated, simultaneously triggering a signal to reset the integrator $\int I{\rm d}t \geq \phi$.
This process is quite similar to the integrate-and-fire neuron.
Distinct luminance stimuli denoted as $I$ result in varying spike firing rates, where the initiation of output and reset operations occurs asynchronously across multiple pixels.
As a general trend, greater light intensity corresponds to higher firing speeds.
The raw data captured by the spike camera takes the form of a three-dimensional spike array denoted as $D$.
The spike camera's primary focus lies in integrating luminance intensity and emitting spikes at an exceptionally high frequency (20,000 Hz).
During each sampling timestep, when a spike has just been discharged, a digital signal of ``1'' (indicating a spike) is produced; otherwise, a signal ``0'' is generated.

\noindent\textbf{Spatio-temporal Analysis on Spike Visual Saliency} 
Saliency object detection (SOD) is a task that segments the regions or objects of greatest interest in human vision from the scene. 
Spike cameras record the scenes through accumulating intensity and generate sparse spike data, the spike visual saliency is closer to the biological principles of human eyes.
In a spike camera, when the firing threshold $\phi$ is reached, the integrator is reset and triggers a spike emission. The time it takes for the integrator to fill from empty to capacity is not fixed due to fluctuations in light conditions.
At a microscopic level, the act of firing a spike corresponds to the recording of a consistent number of photons.
Different from the conventional SOD utilizing standard cameras, the visual saliency within the continuous spike stream is hidden within the binary spikes in the spatio-temporal domain.
As depicted in Fig.~\ref{fig2}, given the binary nature of the spike stream, extracting saliency regions at specific time points necessitates simultaneous consideration of spatial and temporal factors.

\noindent\textbf{Spike-based Visual Saliency Dataset}
\label{sec:dataset}
The datasets play an important role for the development of new algorithms and models. In our paper, we construct the first spike-based visual saliency (SVS) dataset. We use a spike camera (Spatial resolution of 250$\times$400 and a temporal resolution of 20,000 Hz.) to collect real-world spike data, which includes different light intensity scenes. We use the average Light Intensity Scale (LIS) to split the high and low-intensity scenes, the LIS is defined as:
\begin{equation}
    \rm\textbf{LIS} = {\rm\textbf{M}} / ({\rm\textbf{H} \times \rm\textbf{W}})
\end{equation}
where $ \rm\textbf{M} $ is the number of the spike in a frame, $ \rm\textbf{H} $ and $ \rm\textbf{W} $ is the camera size, the details of dataset are listed in Table \ref{tab1}.

\begin{table}[t]
    \footnotesize
    \centering
    \tabcolsep=0.18cm
    \begin{tabular}{lcccccc}
     \toprule[1pt]
         \multirow{2}{*}{} & \multicolumn{2}{c}{Train} & \multicolumn{2}{c}{Val.} & \multirow{2}{*}{Total}  \\
         \cmidrule(r){2-3}  \cmidrule(r){4-5}
         & high & low & high & low  \\
         \midrule
         Seq. num. & 24 & 76 & 8 & 22 & 130 \\
         Spikes num. & 8.7B & 10.3B & 2.9B & 3.1B & 25B\\
         Mean spikes & 0.36B & 0.13B & 0.37B & 0.14B & 0.19B\\
         Mean LIS & 0.045 & 0.017 & 0.046 & 0.018 & 0.031\\
         Mean objs. & 2.6 & 1.8 & 2.9 & 2.2 & 2.1\\
         Mean size (Pixels) & 7891 & 7163 & 4667 & 8964 & 7449\\
     \bottomrule[1pt]
    \end{tabular}
    \caption{{The statistics of SVS dataset.}}
    \label{tab1}
\end{table}

\begin{figure*}[htbp]
    \centering
    \includegraphics[width=1\linewidth]{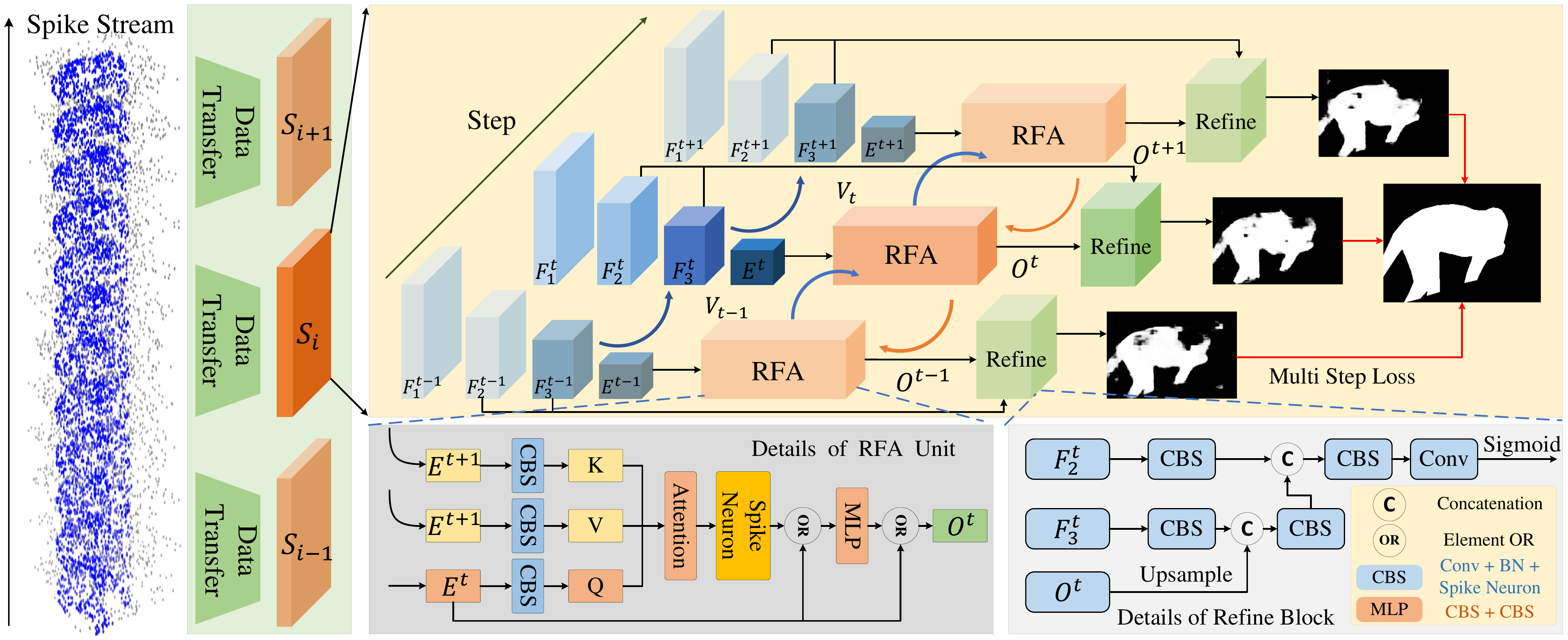}
    \caption{{The framework of our Recurrent Spiking Transformer (RST).} Our recurrent spiking Transformer is a full spiking neural network architecture, which comprises spike-based spatio-temporal feature extraction, recurrent feature aggregation, multi-scale refinement, and multi-step loss.}
    \label{fig4}
\end{figure*}

Our dataset comprises 130 spike sequences, each of which is divided into 200 subsequences. Initially, we employ a spike-based reconstruction method \cite{zhu2019retina} to reconstruct textures, and annotate salient objects with instance labels on them. To facilitate training and evaluation, we partition the dataset into a training set and a validation set. The training set encompasses 100 sequences, encompassing 20,000 annotated frames, while the validation set consists of 30 sequences with 6,000 annotated frames. The annotated frames have a time interval of 20 $ms$, which corresponds to 400 spike frames. For visual reference, example annotations from our dataset can be seen in Fig. \ref{fig3}. Within our dataset, we offer spike frames, reference scenes, and object masks, all of which are accessible to the research community.

\section{Learning to Detect Visual Saliency via Spiking Neural Network}

\noindent\textbf{Preliminary: Spiking Neural Network} 

\noindent\textbf{1) Spiking Neuron.}
Different from traditional ANN models use a weighted sum of inputs to generate continuous values, SNN models transmit discrete spikes by combining the weighted sum of inputs and the membrane potential of the spiking neuron. If the membrane potential reaches a threshold $ \rm\textbf{V}_{th} $, the neuron will emit spike $ \rm\textbf{S}_{t}\in\{0, 1\} $ through a Heaviside step function $ \Theta(\cdot)$ to its subsequent neuron. In this paper, we use Leaky Integrate-and-Fire (LIF) model \cite{gerstner2014neuronal} which is a widely used neuron model in SNN as our basic computing unit, and the dynamics equations of LIF neuron are described as:
\begin{equation}
    \rm\textbf{H}_{t} = \rm\textbf{V}_{t-1} + \frac{1}{\tau} \cdot (\rm\textbf{X}_{t} - (\rm\textbf{V}_{t-1} - \rm\textbf{V}_{reset})),
\end{equation}
\begin{equation}
    \rm\textbf{S}_{t} = \Theta(\rm\textbf{H}_{t} - \rm\textbf{V}_{th}),
\end{equation}
\begin{equation}
    \rm\textbf{V}_{t} = \rm\textbf{H}_{t} \cdot (1-\rm\textbf{S}_{t}) + \rm\textbf{V}_{reset}\cdot \rm\textbf{S}_{t},
\end{equation}
where $\rm\textbf{X}_{t}$ denotes the input to neuron at time t, $ \tau $ is the membrane time constant, $ \rm\textbf{H}_{t} $ is the membrane potential after neuronal dynamics at t and $ \rm\textbf{V}_{t} $ represents the membrane potential after emitting spike.

\noindent\textbf{2) Spiking Transformer Block.}
Spikformer \cite{zhou2022spikformer} introduces the Spiking Self Attention (SSA) in SNN and applies it to the classification task. SSA replaces the nonlinearity function by LIF neuron for each layer to emit spike sequences. Considering the property of SNN, SSA removes the softmax operation for the attention matrix and uses a scaling factor $\rm\textbf{s} $ to constrain the large value of the matrix. Given a feature input $ \rm\textbf{X} \in \mathbb{R}^{B \times N \times C} $, SSA uses learnable matrices $ \rm\textbf{W}_{Q} ,\rm\textbf{W}_{K},\rm\textbf{W}_{V} \in \mathbb{R}^{C \times C} $ and spike neurons $ \rm{SN}_{Q}, \rm{SN}_{K}, \rm{SN}_{V} $ to compute the query ($ \rm\textbf{Q} $), key ($ \rm\textbf{K} $), and value ($ \rm\textbf{V} $):
\begin{equation}
    \begin{split}
        \rm\textbf{Q} &= \rm{SN}_{Q}(\rm{BN}(\rm\textbf{X}W_{Q})), \\ 
        \rm\textbf{K} &= \rm{SN}_{K}(\rm{BN}(\rm\textbf{X}W_{K})), \\
        \rm\textbf{V} &= \rm{SN}_{V}(\rm{BN}(\rm\textbf{X}W_{V})),
    \end{split}
\end{equation}
where $\rm{BN} (\cdot)$ is Batch Normalization operation, and $ \rm\textbf{Q},\rm\textbf{K},\rm\textbf{V} \in \mathbb{R}^{B \times N \times C} $. Then the SSA can be computed as:
\begin{equation}
    \begin{split}
        &\rm\textbf{SSA}'(\rm\textbf{Q},\rm\textbf{K},\rm\textbf{V}) = SN(\rm\textbf{Q}\rm\textbf{K}^{T}\rm\textbf{V} \cdot \rm\textbf{s}), \\
        &\rm\textbf{SSA}(\rm\textbf{Q},\rm\textbf{K},\rm\textbf{V}) = SN(\rm{BN}(Linear(\rm\textbf{SSA}'(\rm\textbf{Q},\rm\textbf{K},\rm\textbf{V})))). \\
    \end{split}
\end{equation}

We notice that Spikformer uses the same operations as the traditional Transformer encoder after computing self-attention. The element-add operation is used between each SSA layer and the output of SSA layer $ \rm\textbf{O} \in \mathbb{N} $, which means that $ \rm\textbf{O} \notin \{0,1\} $ is no longer a binary spike sequence.
In order to solve this problem and adapt SSA for our task, we propose a Recurrent Spiking Transformer (RST) to facilitate complete binary spike communication while enhancing the extraction of temporal information.

\noindent\textbf{Temporal Spike Representation.} 
To effectively leverage the temporal information of the spike stream, we employ the inter-spike interval as the spike representation. The intensity is directly correlated with either the spike count or spike frequency. Consequently, by utilizing the inter-spike intervals or straightforwardly tallying the spikes over a specific period, the temporal information in the scene can be comprehensively represented:
\begin{equation}
    \rm\textbf{S} = C / \Delta t_{x,y},
\end{equation}
where $\rm C$ denotes the maximum grayscale, and ${\rm \Delta t_{x,y}}$ means the spike firing interval at pixel $\rm (x, y)$.

\noindent\textbf{Spike-based Spatio-temporal Feature Extraction.}
Inspired by the temporal property of SNN, we use spike neuron to extract spike-based spatio-temporal feature. The SNN needs recurrent multi-steps to get rich features, so given a temporal spike representation $ \rm\textbf{S} \in \mathbb{R}^{C \times H \times W} $, we first repeat $ \rm\textbf{S} $ for $ \rm\textbf{T} $ steps as $ \rm\textbf{S}' \in \mathbb{R}^{T\times C \times H \times W} $ and use a CBS module (i.e., Conv + BN + Spiking Neuron) to generate multi-scale feature for each step parallelly. The CBS module is consist with a 2D convolution layer (stride 1, kernel size 3), a BatchNorm layer, a LIF neuron and a max pooling layer (stride 2):
\begin{equation}
    \rm\textbf{F} = MP(SN(BN(Conv2d(\rm\textbf{S}')))).
\end{equation}

Similar to traditional SOD model, we use 4-block module to extract feature $ \rm\textbf{F}_{i} \in \mathbb{R}^{T \times \frac{D}{2^{4-i}} \times \frac{H}{2^i}  \times \frac{W}{2^i}} $, where $\rm i\in [1,4]$ and the $ \rm\textbf{D} $ is the dimension of Recurrent Spiking Transformer (RST). Vanilla Vision Transformer \cite{dosovitskiy2020image} usually add position embeddings for image patches, but the spike-based feature has a natural representation for the salient area, so we just use the identity feature and flatten the last feature $ \rm\textbf{F}_{4} \in \mathbb{R}^{T \times D \times \frac{H}{16} \times \frac{W}{16}} $ as the input $ \rm\textbf{E} \in \mathbb{R}^{T \times D \times N } $ of our RST module. It not only uses the property of spike-based feature, but also keeps the spiking propagation between each module.

\noindent\textbf{Recurrent Feature Aggregation (RFA) via Spiking Transformer.}
\begin{figure}[t]
    \centering
    \includegraphics[width=\linewidth]{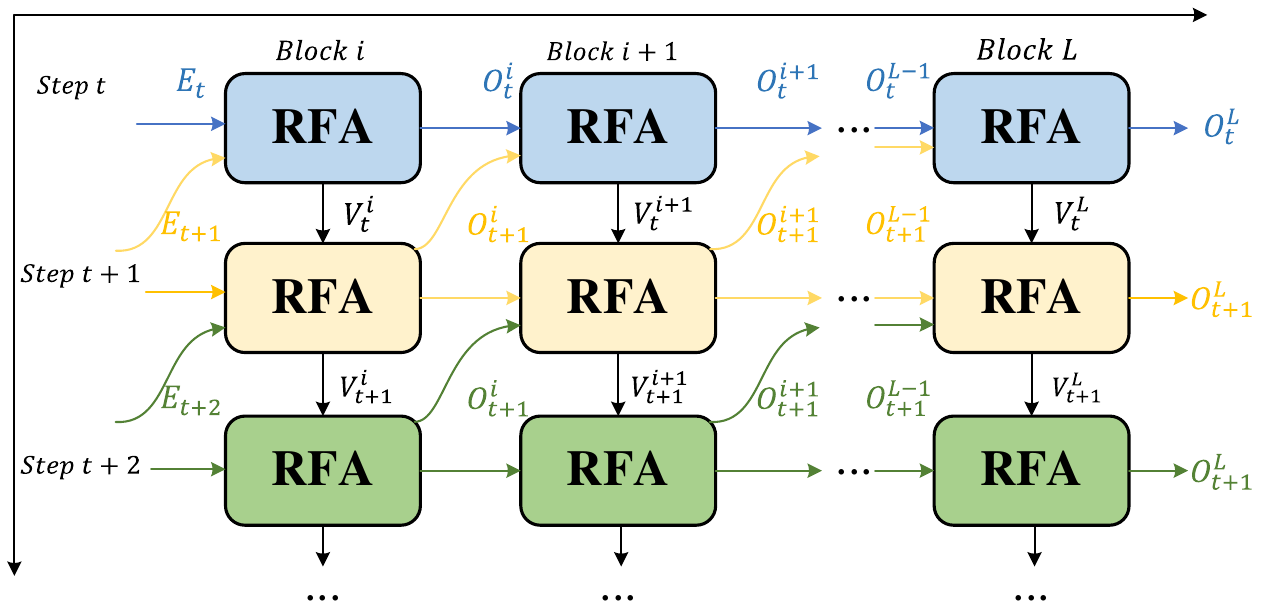}
    \caption{{Recurrent mode of our RFA module.} RFA uses attention mechanism to aggregate the adjacent step features $ \rm{E}_{t} $ and $ \rm{E}_{t+1} $, which will enhance the feature and generate a better saliency map at the step $\rm{t}$. }
    \label{fig5}
\end{figure}
\begin{figure*}[t]
    \centering
    \includegraphics[width=1\linewidth]{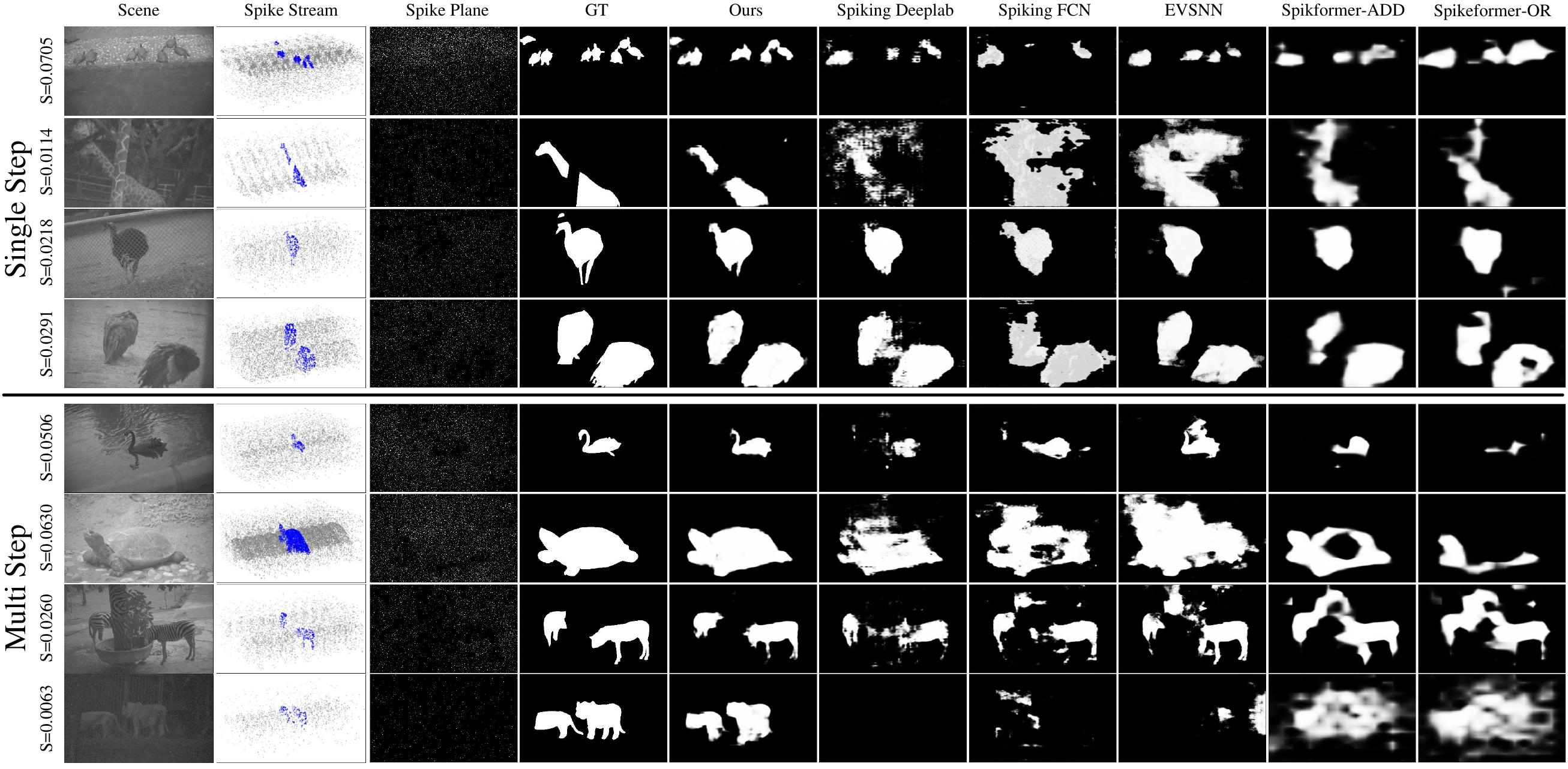}
    \caption{{Qualitative results on our SVS dataset.} $S$ denotes the light intensity scale of the scene. Spikformer-ADD employs non-spikes in its residual connection, while the remaining methods utilize a full spiking neural network architecture. Our model excels in capturing finer details compared to other SNN-based methods in both single-step and multi-step settings.}
    \label{fig7}
\end{figure*}
The temporal property of SNN is dependent on the accumulation of membrane potential $\rm\textbf{V}$, only use $ \rm\textbf{V}_{t-1} $ to generate $ \rm\textbf{E}_{t} $ may get sparse feature at the early step. Because we parallelly extract features for all steps, the feature $ \rm\textbf{E} $ can be split as $ [\rm\textbf{E}_{1},\rm\textbf{E}_{2},...,\rm\textbf{E}_{T}] $ and $ \rm\textbf{E}_{t} \in \mathbb{R}^{1 \times D \times N} $ is the feature at step $ \rm\textbf{t} $, so we use $ \rm\textbf{E}_{t+1} $ to enhance current feature. At step $ \rm\textbf{t} $, the recurrent spiking transformer block receives $ \rm\textbf{E}_{t} $ as Query branch and $ \rm\textbf{E}_{t+1} $ as Key and Value branch to calculate $ \rm\textbf{Q}_{t},\rm\textbf{K}_{t},\rm\textbf{V}_{t} \in\mathbb{R}^{1 \times D \times N} $:
\begin{equation}
    \begin{split}
        \rm\textbf{Q}_{t} &= \rm{SN}_{Q}(\rm{BN}(\rm\textbf{E}_{t}W_{Q})),\\ 
        \rm\textbf{K}_{t} &= \rm{SN}_{K}(\rm{BN}(\rm\textbf{E}_{t+1}W_{K})),\\ 
        \rm\textbf{V}_{t} &= \rm{SN}_{V}(\rm{BN}(\rm\textbf{E}_{t+1}W_{V})).\\ 
    \end{split}
\end{equation}

\begin{table*}[t]
  \small
  \centering
  \tabcolsep=0.43cm
  \begin{center}
    \begin{tabular}{lcccccccccccc}
    \toprule[1pt]
     \multirow{2}{*}{Method} & \multicolumn{4}{c}{Single Step} &  \multicolumn{4}{c}{Multi Step} \\

        \cmidrule(r){2-5} \cmidrule(r){6-9}
        
        & MAE$\downarrow$ & ${F}_{\beta}^{max}\uparrow$ & ${mF}_{\beta}\uparrow$ & ${S}_{m}\uparrow$  &  MAE$\downarrow$ & ${F}_{\beta}^{max}\uparrow$ & ${mF}_{\beta}\uparrow$ & ${S}_{m}\uparrow$\\ 
    
    \midrule
    
     Spiking Deeplab & 0.1026 & 0.5310 & 0.5151 & 0.6599 & 0.0726 & 0.6175 & 0.6051 & 0.7125 \\
     Spiking FCN & 0.1210 & 0.4779 & 0.4370 & 0.6070 & 0.0860 & 0.5970 & 0.5799 & 0.6911 \\
     
     EVSNN & 0.1059 & 0.5221 & 0.4988 & 0.6583 & 0.0945 & 0.6267 & 0.5850 & 0.7023  \\
     Spikformer-ADD & 0.1185 & 0.4638 & 0.4415 & 0.6119 & 0.0717 & 0.6890 & 0.6731 & 0.7563 \\

     Spikformer-OR & 0.1389 & 0.4527 & 0.4408 & 0.6068 & 0.0738 & 0.6526 & 0.6323 & 0.7161 \\
     
     \midrule
     
     \textbf{Ours} & \textbf{0.0784} & \textbf{0.6313} & \textbf{0.6171} & \textbf{0.6970} & \textbf{0.0554} & \textbf{0.6981} & \textbf{0.6882} & \textbf{0.7591} \\
    \bottomrule[1pt]
    
    \end{tabular}
  \end{center}
  \caption{{Quantitative comparison on our SVS dataset.} Spikformer-ADD employs non-spikes in its residual connection, while the remaining methods utilize a full spiking neural network architecture.}
  \label{table2}
\end{table*}

Then we reshape the features as $ \rm\textbf{Q}_{t}',\rm\textbf{K}_{t}',\rm\textbf{V}_{t}' \in\mathbb{R}^{1 \times n \times N \times \frac{D}{n}} $ and calculate multi-head attention between adjacent step:
\begin{equation}
    \rm\textbf{AQ}_{t}' = SN(\rm\textbf{Q}_{t}' \rm\textbf{K}_{t}'^{T} \rm\textbf{V}_{t}' \cdot \rm\textbf{s}),
\end{equation}
where $ \rm\textbf{n} $ is the number of multi head attention, $ \rm\textbf{s}=\sqrt{\frac{n}{D}} $, and $\rm\textbf{AQ}_{t}'$ will be reshaped as $\rm\textbf{AQ}_{t} \in \mathbb{R}^{1\times D \times N}$. Then we use a Linear layer to project the feature and a residual element-or connection to select the salient area:
\begin{equation}
    \rm\textbf{Z}_{t} = \rm\textbf{E}_{t} \lor SN(BN(Linear(\rm\textbf{AQ}_{t}))).
\end{equation}

The feature $\rm\textbf{Z}_{t}$ will be sent to an MLP module which consists of two CBS blocks to get the output $\rm\textbf{O}_{t} \in \mathbb{R}^{1\times D \times N}$:
\begin{equation}
    \rm\textbf{O}_{t} = \rm\textbf{E}_{t} \lor MLP(\rm\textbf{Z}_{t}).
\end{equation}

Finally, we parallel apply this operation for each step feature. As for the final $ \rm\textbf{T}$ step, use its identity feature for each branch. After that, we concat $[\rm\textbf{O}_{1},\rm\textbf{O}_{2},...,\rm\textbf{O}_{T}]$ as $\rm\textbf{O} \in \mathbb{R}^{T\times D \times N}$, reshape it to $ \rm\textbf{F}' \in \mathbb{R}^{T \times D \times \frac{H}{16} \times \frac{W}{16}}$ and feed it to Multi-scale Refinement Block.

\noindent\textbf{Multi-scale Refinement.}
Different from the SNN-based classification task, the saliency map is closely related to the feature spatial size and the semantic information, so it is necessary to have a multi-scale refinement module. In this section, we design a Spiking Multi-scale Refinement block to aggregate the semantic information from different scale features. The refinement block uses CBS block as the basic unit and nearest interpolation to upsample feature. In our model, $ \rm\textbf{F}_{2}, \rm\textbf{F}_{3}, \rm\textbf{F}' $ are used for feature refinement and upsample, and the output $ \rm\textbf{S} \in \mathbb{R}^{T \times D \times \frac{H}{4} \times \frac{W}{4}} $ will forward a $ 1\times1 $ Conv2d layer and a Sigmoid function to generate the saliency map.

\noindent\textbf{Efficient Multi-step Loss.}
Traditional SNN-based methods for segmentation and classification tasks usually calculate average value for multi-steps as the final result, it may lose information along the time dimension in our task. To better use the relationship among multi-steps, we respectively calculate the loss for every step result, it can also establish constraints for early step's feature. We use binary cross entropy loss $ \mathcal{L}_{bce} $ \cite{de2005tutorial}, IoU Loss $ \mathcal{L}_{iou} $ \cite{rahman2016optimizing} and SSIM Loss $ \mathcal{L}_{ssim} $ \cite{wang2004image} to train our model. And for a SNN model with step $ \rm\textbf{T} $, the final loss $ \mathcal{L} $ can be expressed as:
\begin{equation}
    \mathcal{L} = \sum_{i=1}^{\rm\textbf{T}} {\alpha}_{i}( \mathcal{L}_{bce} + \mathcal{L}_{iou} + \mathcal{L}_{ssim} ),
\end{equation}
where ${\alpha}_{i} = ({\rm\textbf{T}}-i+1) / {\sum_{i=1}^{\rm\textbf{T}}}$ is the weight of each step, we set $\rm\textbf{T}=5$ in our experiment.

\section{Experiment}
\subsection{Experiment Setup}
\noindent\textbf{Dataset.}
We use our spike-based visual saliency (SVS) dataset to test and verify our proposed method. The details are shown in Table \ref{tab1}.

\noindent\textbf{Comparative SNN Models.}
Since there is no SNN-based salient object detection method, we compare our methods with four mainstream SNN-based architectures. Spiking Deeplab and Spiking FCN \cite{kim2022beyond} are methods for semantic segmentation, Spikformer \cite{zhou2022spikformer} is designed for classification tasks, and the EVSNN \cite{zhu2022ultra} uses SNN for event-based video reconstruction. We modify these methods to adapt spiking-based SOD and train them on SVS dataset. 

\noindent\textbf{Training Details.}
For a fair comparison, we use the same setting for all methods. 
AdamW is used to train 20 epochs for all models and the initial learning rate is set to $2\times10^{-5}$, which linearly decays with the epoch until $2\times10^{-6}$. 
We use $ 256\times256 $ as the input size and the time interval of spike data is set to 0.02s, which means the methods will get 400 frames spike at each iteration. We respectively train the model on two settings: single-step and multi-step. When using multi-step mode for training, the same spike data is input at each step and the model iterates five steps, which will result in better performance on a single frame. When using single-step mode, we input continuous spike data in the temporal domain, and the model only iterates once for each input.

\noindent\textbf{Evaluation Metrics.}
Inspired by traditional SOD tasks, we use Mean absolute error (MAE) \cite{borji2015salient}, maximum F-measure ${F}_{\beta}^{max}$, mean F-measure ${mF}_{\beta}$ \cite{achanta2009frequency}, and Structure-measure ${S}_{m}$ \cite{fan2017structure} as our evaluation metrics, to evaluate the quality between predict saliency map $\rm\textbf{S}$ and ground-truth label $\rm\textbf{G}$.

\subsection{Quantitative Experiment}
Table \ref{table2} shows the quantitative results for all methods using different steps on our SVS dataset, and our method has the best performance on both two settings. Notice that the step setting has a significant influence on all methods, the reason is that spiking neurons need some steps to accumulate the membrane potential. The Spikformer-ADD model has a better result on multi-steps than other comparison methods, this is because Spikformer-ADD uses the element-add operation for residual connection in its SSA module to enhance the feature that transfers floating numbers between each block. If we replace the element-add as the element-or operation, the Spikformer-OR has lower performance. Although our model transfers the whole spike-based features among all modules, our method can predict better than Spikformer-ADD.

\subsection{Qualitative Experiment}
Fig. \ref{fig7} illustrates the results of various methods in both single and multi-step modes. Notably, when confronted with intricate scenes featuring comparable objects and backgrounds, our method excels in delineating object contours and edges, surpassing other approaches. Furthermore, our method exhibits remarkable robustness across diverse illumination conditions, generating distinct saliency maps for target objects even in challenging low-light scenes, unlike other comparison methods that experience diminished effectiveness in such scenarios.

\subsection{Performance Analysis in the Temporal Domain}
\begin{figure}[b]
    \centering
    \includegraphics[width=\linewidth]{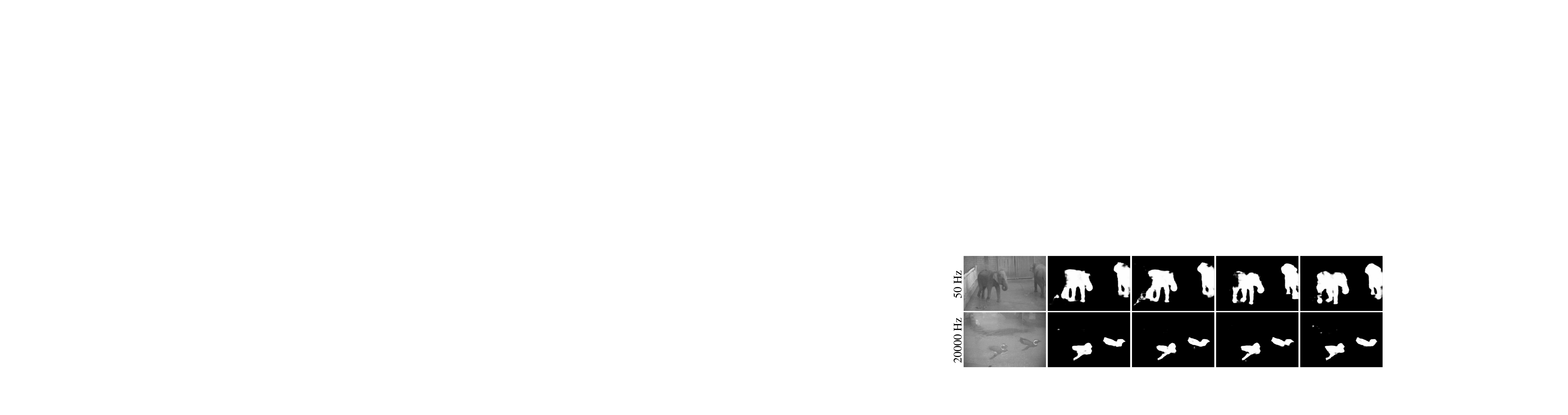}
    \caption{{Results from our model during single-step inference using continuous spike data input.} The top row illustrates results from 50 Hz spike data input, while the bottom row showcases results from 20,000 Hz spike data input.}
    \label{fig8}
\end{figure}

Benefiting our recurrent spiking Transformer module, our model can be easily extended to continuous salient object detection. Unlike other SNN-based methods that necessitate multiple steps for sufficient information extraction, our model achieves high-quality prediction results with just a single inference step.
The continuous detection results are depicted in Fig. \ref{fig8}. Remarkably, our model accurately predicts results even with a 20,000 Hz spike input, where each input corresponds to a single step. This remarkable efficiency leads to minimal energy consumption during continuous spike stream processing.
In direct comparison with its ANN-based counterpart, which consumes 167 mJ per inference, our method operates at a mere 5.8 mJ, signifying a substantial reduction in power usage by a factor of 28.7. Further details are available in our supplementary materials.

\begin{table}[t]
\small
    \centering
    \tabcolsep=0.23cm
    \begin{tabular}{lccccc}
    \toprule
        Recurrent Mode &  MAE$\downarrow$ & ${F}_{\beta}^{max}\uparrow$ & ${mF}_{\beta}\uparrow$ & ${S}_{m}\uparrow$ \\
        \midrule
        Vanilla & 0.0581 & 0.6811 & 0.6716 & 0.7522 \\
        Forward & 0.0611 & 0.6696 & 0.6599 & 0.7432 \\
        Ours (Reverse) & \textbf{0.0554} & \textbf{0.6981} & \textbf{0.6882} & \textbf{0.7591} \\
    \bottomrule
    \end{tabular}
    \caption{{Effect of different recurrent modes.}}
    \label{tab:table3}
\end{table}
\begin{table}[t]
\small
    \centering
    \tabcolsep=0.19cm
    \begin{tabular}{lccccc}
    \toprule     
        Method  &  RFAs  & MAE$\downarrow$ & ${F}_{\beta}^{max}\uparrow$ & ${mF}_{\beta}\uparrow$ & ${S}_{m}\uparrow$ \\
        \midrule
        
        w/o Refine & 6 & 0.0701 & 0.6298 & 0.6205 & 0.7190 \\

        \midrule

        Refine & 0 & 0.0642 & 0.6789 & 0.6622 & 0.7385 \\
        Refine & 2 & 0.0571 & 0.6942 & 0.6829 & 0.7548 \\
        Refine & 4 & 0.0559 & 0.6965 & 0.6848 & 0.7563 \\
        Refine & 8 & 0.0580 & 0.6818 & 0.6716 & 0.7466 \\
        
        Ours & 6 & \textbf{0.0554} & \textbf{0.6981} & \textbf{0.6882} & \textbf{0.7591} \\
    \bottomrule
    \end{tabular}
    \caption{{Effect of RST module and refine module.}}
    \label{tab:table4}
\end{table}
\begin{table}[t]
\small
    \centering
    \begin{tabular}{lccccc}
    \toprule
        Recurrent Mode & Loss &  MAE$\downarrow$ & ${mF}_{\beta}\uparrow$ & ${S}_{m}\uparrow$ \\
        \midrule
        
        Vanilla & Vanilla & 0.0635 & 0.6690 & 0.7521 \\
        Vanilla & Multi step & 0.0581 & 0.6716 & 0.7522 \\
        
        Reverse & Vanilla & 0.0613 & 0.6872 & \textbf{0.7666} \\
        Reverse & Multi step & \textbf{0.0554} & \textbf{0.6882} & 0.7591 \\
    \bottomrule
    \end{tabular}
    \caption{{Effect of the multi-step loss.}}
    \label{tab:table5}
\end{table}
\begin{table}[htbp]
\small
    \centering
    \tabcolsep=0.31cm
    \begin{tabular}{lccccc}
    \toprule
        Operation &  MAE$\downarrow$ & ${F}_{\beta}^{max}\uparrow$ & ${mF}_{\beta}\uparrow$ & ${S}_{m}\uparrow$ \\
        \midrule
        ADD & 0.0576 & 0.6803 & 0.6710 & 0.7540 \\
        Concat & 0.0567 & 0.6933 & 0.6834 & 0.7563 \\
        Ours (OR) & \textbf{0.0554} & \textbf{0.6981} & \textbf{0.6882} & \textbf{0.7591} \\
    \bottomrule
    \end{tabular}
    \caption{{Effect of the element-wise operation in RST.}}
    \label{tab:table6}
\end{table}

\subsection{Ablation Study}
\noindent\textbf{Effect of Recurrent Spiking Transformer.}
In spiking neurons, the membrane potential is useful for extracting spatio-temporal features. Maximizing the utility of these features across all steps promises enhanced model performance compared to the vanilla SNN propagation mode. As shown in Table \ref{tab:table3}, we test the effect of different recurrent modes. The ``{Vanilla}'' means the SNN architecture without additional recurrent structure, ``Forward'' means using the output of step ${t-1}$ as the key and value batch of RST module. ``Reverse'' is our recurrent mode shown in Fig.~\ref{fig5}. ``Forward'' get a worse result than ``Vanilla'', this can be attributed to sparse information in the early steps, potentially leading to unfavorable effects when directly fusing features.
``Reverse'' mode operates in parallel, efficiently enhancing features by fusing those from step ${t+1}$, thus showcasing its effectiveness in bolstering current features.

\noindent\textbf{Effect of RFA and Refine Module.}
As shown in Table \ref{tab:table4}, we test the effect of the number of RFA modules and the refine block.
Removing the refine block results in a significant performance drop, emphasizing its necessity for robust dense pixel prediction tasks. The influence of RST modules on the final results is evident, yet a noteworthy observation is that performance improvement does not exhibit a linear trend with an increasing number of RST modules. This could be attributed to challenges in effectively training larger models within the limitations of the dataset.

\noindent\textbf{Effect of Multi-step Loss.}
We compare the effect of the vanilla loss and our multi-step loss, the results are shown in Table \ref{tab:table5}. 
Our multi-step loss assigns greater weight to early step results, aiding the model in concentrating on sparse features at the beginning. This strategic approach mitigates SNN's reliance on step size to a certain degree, ultimately reducing prediction error rates.

\noindent\textbf{Element-Wise Operation in RST.}
We test the effect of element-wise operation in our RST module. As shown in Table \ref{tab:table6}, the ``{element-and}'' operation can cause all spikes to 0 on the training stage, so it is difficult to train this model. The ``{Concat}'' operation can get similar results to our method, but the computation complexity will increase rapidly as the dimension rises. The ``{ADD}'' operation performs worse than others, the reason is that usage of the refine block will convert the features back to spike-based binary features. Considering both the energy consumption and the performance, we use ``{OR}'' operation in our model.

\section{Conclusion}
In this paper, we explore visual saliency in continuous spike streams using the spike camera. We introduce the Recurrent Spiking Transformer framework, efficiently extracting spatio-temporal features for visual saliency detection while minimizing power consumption. Our constructed spike-based dataset validates the superiority of our RST framework over other SNN-based models, advancing spike-based saliency detection and offering a fresh perspective for SNN-based transformers. This study also innovates transformer models in continuous spike stream analysis.

\section{Acknowledgments}
This work is partially supported by grants from the National Natural Science Foundation of China under contract No. 62302041 and China National Postdoctoral Program under contract No. BX20230469.

\bibliography{aaai24}

\end{document}